\documentclass{llncs}
\usepackage{epsf}
\usepackage{graphicx,booktabs}
\graphicspath{ {images/} }
\usepackage{cleveref}
\usepackage{tabularx}
\usepackage{mathtools}
\usepackage{rotating}
\usepackage{multirow}
\usepackage{caption}
\usepackage{enumitem}
\setlist[itemize]{noitemsep, topsep=0pt}
\begin{document}

\title{Experiments~with~Neural~Networks~for
Small~and~Large~Scale~Authorship~Verification}

\author{
Marjan Hosseinia and Arjun Mukherjee}
\institute{University of Houston\\
Department of Computer Science\\
           mhosseinia@uh.edu, arjun@cs.uh.edu
}

\maketitle

\begin{abstract}
We propose two models for a special case of authorship verification problem. The task is to investigate whether the two documents of a given pair are written by the same author. We consider the authorship verification problem for both small and large scale datasets. The underlying small-scale problem has two main challenges: First, the authors of the documents are unknown to us because no previous writing samples are available. Second, the two documents are short (a few hundred to a few thousand words) and may differ considerably in the genre and/or topic. To solve it we propose transformation encoder to transform one document of the pair into the other. This document transformation generates a loss which is used as a recognizable feature to verify if the authors of the pair are identical. For the large scale problem where various authors are engaged and more examples are available with larger length, a parallel recurrent neural network is proposed. It compares the language models of the two documents. We evaluate our methods on various types of datasets including Authorship Identification datasets of PAN competition, Amazon reviews and machine learning articles. Experiments show that both methods achieve stable and competitive performance compared to the baselines.
\end{abstract}

\section{Introduction}
In this paper, we consider the problem of Authorship Verification (AV) which is a branch of forensic authorship analysis. When given two text documents, we look to verify whether the two documents are written by the same author while no previous writing samples of their author/authors have been specified.\par 
Majority of online services work on textual communications between users. Their overall reliability and performance can be impacted by someone who abuses the application and provides scripts while hiding their real identity and pretending to be someone else. To preserve the reliability  of such services, the identity of the users should be monitored based on their provided scripts. The authorship verification techniques match the identity of the users with their writing styles. Indeed, authorship verification has an important impact on online document analysis such as plagiarism analysis, sockpuppet detection, blackmailing and email spoofing prevention, to name a few \cite{hosseinia2017detecting}.\par
Traditionally, the studies on AV problem considered a closed and limited set of authors and a closed set of documents written by those authors. During the training step some of these documents (which were sometimes as long as a whole novel) were observed. Then, the problem was to identify whether the authors of a pair of the documents from the rest of the document set were identical~\cite{koppel2004authorship,manevitz2007one,japkowicz1995novelty}. This type of AV problem  benefits from having access to the  writing samples of \textit{future authors} during the training step which is not always realistic. Actually, this structure is  static and is not compatible with new future unseen authors. Recently, the structure of AV problem has changed and became more challenging. Based on the new structure, we are given some document pairs with their binary authorship status. The same-authorship status indicates that both are written by one author while the different-authorship status shows the pairs are written by two individual authors. Based on this binary structure the goal is no longer to learn the writing style of each underlying authors individually (like in the traditional AV methods) but is to learn the difference or similarity of the writing styles of the two types of document pairs.\par
In this paper we define two different schemas to study the AV problem. Under the first schema we address the following challenges: 1- writing samples of available authors are quite limited during the training step as the length of the given text documents is short (a few hundred to a few thousand words) and size of the training set is so small (from  10 to 200 examples). So, it is quite hard to infer the same or different-authorship status of  given  pairs.
2- The test and train documents are from different genera and/or topics which makes the learning and  prediction process much harder as the word distribution might differ considerably.
3- No writing samples of the future authors is specified to us during the training and we may have seen no samples by the future authors at all.
Under the second schema the scale of the training data is larger compared to the first schema. However, we address the problem of identifying the difference in documents from identical domains in two ways: 1- authorship \textit{diversity} in \textit{similar} contents by utilizing Amazon reviews from $300$ distinct authors. 2- Scientific documents from the same area of research by different authors who have almost identical level of expertise in the field. It also can be considered as an application of plagiarism detection.\par We analyze authorship verification on several datasets with binary structure. To our knowledge this amount of analysis has not been done in authorship verification on diverse types of datasets. Two models are proposed. First, a Transformation Encoder (TE) to model error feature vectors for classification inspired by the idea of autoencoders. TE is compatible with the AV problems with small-scale training sets. Giving a pair of input documents, TE transforms one input into the other. In this process, the transformation loss is observed as a reasonable measure of closeness of the two inputs to be used by a classifier. The second model is a parallel recurrent neural network (PRNN) that is inspired by the popular similarity measures in Statistical Machine Learning (ML). Being based on language models, it is  mostly applicable for relatively larger datasets. PRNN compares the proximity of the language model of its two input sequences to investigate their authorship.
We also propose the \textit{summary vector} to adapt our problem to a common binary classification style to create strong baselines as there are limited studies in authorship verification according to the literature. Applying this adaptation we are able to employ the recognized classifiers as well as similarity measures that are widely used in ML to build our baselines.
Besides, the two pre-existing datasets, Amazon reviews and MPLA-400, are mapped to the binary structure to be used for our large scale AV problem. \par 
Experimental results on evaluation datasets show that both methods achieve stable and competitive performance compared to the baselines.
\section {System Design} \label{method}
Let $P=(S, T)$ denotes a pair of documents, indicating $S$ as the source and $T$ as the target. Here, the task is to investigate whether $S$ and $T$ are written by the same author. We map this problem into a binary classification paradigm. Accordingly, if $S$ and $T$ are authored by the same person, $P$ belongs to the positive class. Nevertheless ($S$ and $T$ have different authors) $P$ belongs to the negative class. In the first step, we explain the Transformation Encoder which is a feature extraction-based method designed for the small-scale datasets with $200$ labeled samples at most. However, many AV problems might have a larger scale with much more examples. So, we introduce the Parallel Recurrent Neural Network (PRNN) for large scale datasets in the second step.
  
\subsection {Transformation Encoder (TE)} 
\begin{figure}[t!]
\captionsetup{width=0.45\textwidth}
\centering
\begin{minipage}{.5\textwidth}
  \centering
  \includegraphics[scale=0.35]{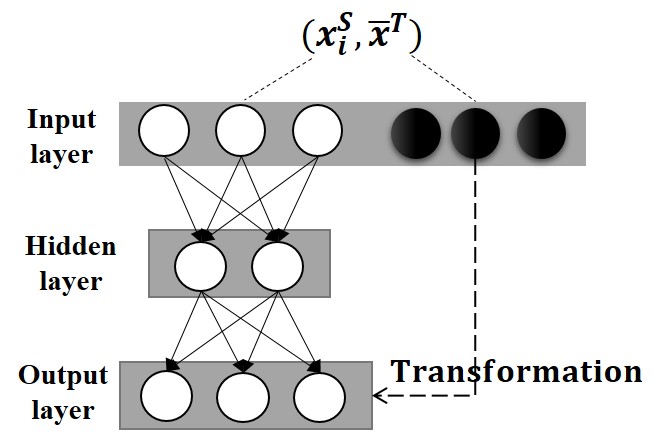}
  \captionof{figure}{Transformation Encoder (TE). $x_i^s$: feature vector of document $i$ after expansion of the source document $S$, $\overline{x}^T$: average feature vector of expanded target document $T$. The dotted line emphasizes that  the reconstructed source should be similar to the target.}
  \label{tae}
\end{minipage}%
\begin{minipage}{.1\textwidth}
\end{minipage}
\begin{minipage}{.45\textwidth}
  \centering
  \includegraphics[scale=0.35]{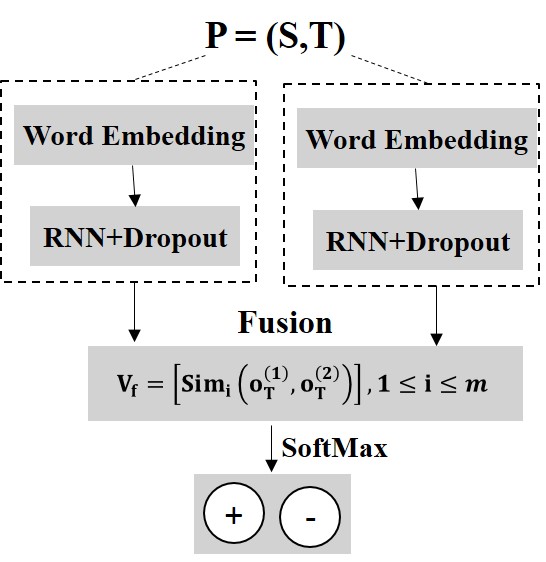}
  \captionof{figure}{PRNN architecture. The network takes two input $S, T$ in parallel and fuses them after passing word embedding and recurrent layers.  }
  \label{prnn arch}
\end{minipage}
\end{figure}

TE is inspired by the idea of an autoencoder. A typical autoencoder neural network is a kind of unsupervised learner exploiting backpropagation in order to reconstruct a given input. In other words, it tries to approximate the identity function. Given an unlabeled input $x \in R^d$, the goal is to learn $W$ and $b$ such that $f_w,_b (x)=x$. It is usually a three-layer neural network, with one input, one hidden and one output layer. $W \in R^{d \times d^\prime}$ is a transformation matrix that encodes a $d$-dimensional input vector $x$ into the mostly lower dimension $d^\prime $. Then, a non-linear function $s$ such as sigmoid will apply to the sum of the transformed vector and a bias $b$ to make the new vector $h$ in the hidden layer.  Finally, the output $z$ will be decoded by applying the same process on $h$ with  $W^\prime \in R^{ d^\prime \times d}$, usually $W^\prime=W^T$, and bias $b^\prime$. Therefore, the goal is to minimize the loss, also known as reconstruction error. To build the Transformation Encoder (Figure \ref{tae}) we add a new input to the structure of autoencoder and modify its reconstruction process. So, TE has two inputs known as source and target. Let $x^s \in R^{d}$ be the first and $x^t \in R^{d}$ be the second input of TE. However, the goal is no longer the reconstruction of one input similar to itself but is to learn a transformation function $g$ that \textit{ reconstructs} the source input $x^s$ similar to the target input $x^t$, i.e., $g_{w,b}(x^s,x^t)=x^t$. Indeed, only the source ($x^s$) passes the reconstruction process (encoding and decoding layers). So, the following steps are kept intact according to a typical autoencoder: $
h^s=s(W^Tx^s+b)  \text{ and } 
z^s=s(W^{\prime}h^s+b\prime)$ where $z^s \in R^{d}$ is the reconstructed input and must be transformed into the target ($z^s \approx x^t$). This can be done by setting TE's objective function as the minimization of the transformation loss. We set the TE transformation loss $Er$ to be the cross-entropy between reconstructed input ($z^s$) and the target input ($x^t$) as: $Er(x^t,z^s)=-\sum_{i=1}^{d}{x^t_i log z^s_i +(1-x^t_i)log(1-z^s_i)}$. \par Now, we assign our authorship verification problem into the proposed Transformation Encoder. It is intuitively expected that TE shows different manner when it transforms the source into the target while both having many features in common compared to the case where they have less common features. Here, the goal is to utilize TE for the AV problems that suffer from restricted labeled data. So,  we put a document expansion method on top of TE as an initial step to overcome the restriction to some extent.
\subsubsection{Document Expansion for Small Scale Datasets }
\label{docexpansion}
Neural networks need sufficient amounts of data during their learning process to avoid the over-fitting problem to produce the desired output. So, we propose a document expansion method to make use of the existing labeled training data of small scale datasets such as PAN to a great extent.
A sliding window with the length of $l$ sentences moves forward through each text document by one sentence per step making a smaller document each time. More specifically, a document with $n$ sentences will be distributed into $n-l+1$ smaller documents. New line characters, as well as empty sentences, are ignored here. So, using this expansion technique each problem $P=(S,T)$ in the small datasets will be converted to $P=(D^S,D^T)$ where $D^S=\{d^s_i \}_{i=1}^{l^S}$ and $D^T=\{d^t_j\}_{j=1}^{l^T}$ are the set of all shorter documents after expansion of $S$ and $T$ (source and target).  $l^S$ and $l^T$ denote the size of $D^S$ and $D^T$ respectively.
\subsubsection{Utilizing TE}
We represent the documents of  $D^S$ and $D^T$ under vector space model. The details are explained in section \ref{setting}. Under one feature set let $x^s_i$ and $x^t_j$ denote the document feature vectors of $d^s_i \in D^S$ and $d^t_j \in D^T $ under vector space model for a problem $P=(D^S,D^T)$  respectively. As mentioned earlier, the goal is to transform the source to the target. However, we have expanded the target document into multiple smaller documents. So, we need to have one single target to be the \textit{shared} target for all the source documents. We compute the average vector of all feature vectors of target documents $d^t_j \in D^T $  as the target input of the TE under vector space model. The average vector is a representative of all features without ignoring any of them despite other techniques such as min/max that only consider the most or least frequent features. The documents are represented under $F$ feature sets separately.
So, under this paradigm all $x^s_i,  1\leq i \leq l^S$ feature vectors  will be transformed into one single target ($\overline{x}^T$). So,  $\overline{x}^T=(1/{l^T})\sum_{j=1}^{l^T} {x^{t}_j}$ where  $x^t_j$ is the feature vector of $jth$ short target document ($d^t_j$) for one problem and $l^t$ is the total number of them. Thus, the final transformation loss of $P=(D^S,D^T)$ is  the average transformation loss of TE across all  $z^s_i, 1\leq i \leq l^S$ (the reconstruction of $x^s_i$) and  $\overline{x}^T$:
$ e_k=(1/{l^S})\sum_{i=1}^{l^S}(Er(\overline{x}^T,z^s_i))$. The value $e_k$ is the  TE error under $k$th feature set  ($1\leq k \leq F$) for a pair of documents $P=(S,T)$ and the critical feature value for a classifier to predict the class label of the given pair. Now, we let $V=[e_k], 1\leq k \leq F$ be the  TE transformation error vector of the problem $P=(S,T)$  where $F$ is the total number of feature sets (at most $7$). The TE error vector $V$ consisting of  transformation errors under different feature sets will be the input of a binary classifier.  These transformation errors are the features to differentiate the pairs in distinct classes. Under this paradigm, the AV problem is mapped to an ordinary binary classification where the TE error vector $V$ of a same-authorship document pair belongs to the positive class and the TE error vector $V$ of a different-authorship document pair is a  member of the negative class.
\subsection{Parallel Recurrent Neural Network (PRNN)}
PRNN is designed to solve the AV problem for relatively large scale datasets. The structure of the problem is the same as TE's. We model a pair of documents using a simple parallel recurrent architecture. The overall model is shown in Figure \ref{prnn arch}. In general, PRNN consists of three components:  two parallel columns of identical layers, one shared fusion layer and a SoftMax layer as the output. We proceed to describe the network in the following paragraphs.
\subsubsection{Parallel columns of embedding and recurrent layers}
Given two documents, the network takes each document as the input of one of the parallel columns separately. In each column, the network embeds all words of the input document through an embedding matrix $E \in R^{d_E \times V_E}$ where $V_E$ is the size of the vocabulary and $d_E$ is the embedding dimension. Then, a fully-connected RNN where the output is to be fed back to its input takes the embedding matrix of the previous layer.
The RNN layer of our model at time $t \in [0,\tau]$ includes:
$h_t=tanh(W^{hh}h_{t-1}+W^{hx}x_{t}+b);$
$o_t=c+W^{ho}h_{t};$ where $x_t$ is the word embedding vector; $b$ and $c$ are the bias vectors; $W^{hh}$, $W^{hx}$ and $W^{ho}$ are the weight matrices. We only take $o_{\tau}$, the output at the last time step $\tau$, as the output of the recurrent layer. Finally, to avoid over-fitting problem we apply dropout regularization to the output of the recurrent layer. It helps the  network to generalize the learnt language models. 
\subsubsection{Fusion layer} 
Let $o_{\tau}^{S}\text{ and } o_{\tau}^{T}$ be the output of the final (RNN) layers after dropout of the two parallel columns. We add a \textit{shared} fusion layer to fuse $o_{\tau}^{S}\text{ and } o_{\tau}^{T}$  by computing several popular similarity measures between them. The resulting fusion vector, $V_f$, is computed as:
$V_{f}=[sim_1(o_{\tau}^S,o_{\tau}^T),..., sim_M(o_{\tau}^S,o_{\tau}^T)]$, $ V \in R^{M}$ where each $sim_{i, 1 \leq i \leq M}$ function belongs to one of the $M$ functions in Table \ref{svector}. Finally, the output layer classifies the fusion vector using a SoftMax function.

\section{Experiment Design}
As the two proposed methods uses different techniques for verification, our experiments to evaluate them are done in two distinct settings: 1-TE schema for Transformation Encoder method, 2-PRNN schema for Parallel RNN.

\begin{table}[t!]
    \begin{minipage}{.5\linewidth}
      \centering       
        \begin{tabular}{lll}
    \toprule
    \textbf{Dataset} & \textbf{Train } &\textbf{Test }  \\
    \midrule
    PAN2013 & 10    & 30     \\
    PAN2014E & 200   & 200    \\
    PAN2014N & 100   & 200    \\
    PAN2015 & 100   & 500     \\
    \midrule
    \textbf{Dataset} & \textbf{Positive } &\textbf{Negative }  \\
     \midrule
    Amazon & 4500   & 4500     \\
    MPLA* & 720   & 720     \\
    \bottomrule
    \end{tabular}%
    \caption{Datasets information}
    \label{panwinners}%
    \end{minipage}%
    \begin{minipage}{.5\linewidth}
    \small    
         \begin{tabular}{ll}
    \toprule
    \textbf{Metric} & \textbf{Description}   \\
    \midrule
    Chi2 kernel &   $\exp(- \gamma \sum_i[\frac{ (x_i - y_i)^2}{ (x_i + y_i)} ])$    \\ Cosine similarity & $xy^T / (||x||||y||)$ \\
    Euclidean  &   $\sqrt{\sum_i{(x_i-y_i)^2}}$    \\ Linear kernel & $x^Ty$  \\
    RBF kernel &$\exp(-\gamma ||x-y||^2)$       \\ Mean of L1 norm & $\sum_i^n|x_i-y_i|/n$ \\
    Sigmoid kernel &   $\tanh(\gamma x^T y + c_0)$ \\     
    \bottomrule
    \end{tabular}%
    \caption{ Similarity functions. $x,y$: document feature vectors, $n$: number of features in $x$ and $y$}
    \label{svector}
    \end{minipage} 
\end{table}

\subsection{Dataset}

\textbf{TE schema} To evaluate the Transformation Encoder we use all available  authorship identification datasets released by PAN \footnote{http://pan.webis.de/data.html} (Table \ref{panwinners}). Each PAN dataset consists of a training and test corpus and each corpus has a various number of distinct problems. One problem is a pair of two documents: the first document of a problem composed of up to five writings (even as few as one) by a single person (implicitly disjoint For  PAN2014 and PAN2015 and explicitly disjoint for PAN2013), and literally the second document includes one piece of writing.  Two documents of a pair might be from significantly various genres and topics. The length of a document changes from a few hundred to a few thousand words. PAN2014 includes two datasets: Essays and Novels.  The paired documents in PAN datasets are used for our experiments. So, for a problem $P=(S, T)$,  $S$  (source) is the first document and $T$ (target) is the second document of a PAN problem.\\
\textbf{PRNN schema} We evaluate PRNN on new schemas of MPLA-400 \footnote{https://github.com/dainis-boumber/MLP-400-datasets} and Amazon reviews. The schemas are defined similarly to the PAN-style explained above. MPLA-400 dataset contains 20 articles by each of the top-20 authors by citation in Machine Learning. We create its new schema, MPLA*, by selecting publications from MPLA-400 that are written by a single author and have no co-authors.  To keep the distribution of authors and classes balanced in MPLA*, we select an equal number of single-authorship articles from all existing 20 authors and map it to the PAN-style (there are at most 9 single-author publications by each author).  Here, the positive class consists of the pairs which are made up of all possible combinations of same-authorship articles ($20 \times \binom {9}{2}=720$). And the same size negative class includes the pairs that are randomly selected from the set of all unique combinations of different-authorship articles. 
We apply the similar method to Amazon review dataset to define its new PAN-style schema.  We select $300$ authors with at least $40$ reviews to make the positive and negative candidate sets. Then, for each author, the positive candidate set is all possible and unique combinations of the author's reviews. To make a positive class we choose 4500 review pairs from this positive candidate set at random. On the other hand, the negative candidate set is made of all unique and possible combinations of review pairs having different authors. The negative class having equal size with the positive class is created by random selection from the negative candidate set.
  
\subsection{Feature Sets and Experiment Settings} \label{setting}
\textbf{TE schema} All documents of $D^S$ and $D^T$ are represented in vector space model under several feature sets with term frequency and boolean feature value assignment separately. Seven feature sets are used: 1-unigram, 2-bigram, 3-trigram, 4-four-gram, 5-unigram Part Of Speech (POS), 6-bigram  POS, 7-char-4gram \footnote{ we use scikit-learn software for all linguistic features}. Note that for each transformation using TE $x^s_i$ and $x^s_j$ are represented under one feature set with one feature value assignment. TE is implemented based on Theano \footnote{http://deeplearning.net/software/theano/}.   The size of the hidden layer is reduced to 50\% of the input layer size with batch size=1 and learning rate=0.1. All classifiers and metrics are implemented by scikit-learn library \cite{pedregosa2011scikit}. Gussian distribution is chosen for Naive Bayes. For K-Nearest Neighbor we set K=3. The L-2 regularization is used for Logistic Regression. For document expansion, we set the size of the sliding window to $l=10$. On average it expands one document into  $30$ smaller documents for  PAN datasets. All other parameters are selected based on pilot experiments. We report accuracy, the Area Under Receiver Operating Characteristic (ROC) curve \cite{egan1975signal} (AUC)  and Score$=$AUC$\times$ Acc in TE experiments. The higher AUC and Score indicate more effective classification.\\
\textbf{PRNN schema} The plain text of each document is used as the input of PRNN. The features sets for the baselines are the same as the TE baselines. However, we did not use the original training and test sets of the PAN datasets as the size of the training set is too small to be used for PRNN. To avoid overfitting problem we perform 5-fold Cross Validation (CV) for the PAN2015, Amazon and MPLA* where we have sufficient amount of examples in training folds. And for the  PAN2013, PAN2014E and PAN2014N datasets that are relatively smaller we perform 10-fold CV to increase the size of the training folds. This setting is applied for PRNN as well as the baselines. We use Theano to implement PRNN. All  classifier's parameters are the same as the TE schema. The back-propagation is done using stochastic gradient descent with learning rate=0.001, batch size=1, and dropout rate=0.2. We use the Glove pre-trained vectors\footnote{https://nlp.stanford.edu/projects/glove/} as an initial value for the embedding vectors when there is a match. Otherwise, a random vector from a continuous uniform distribution over $[0,1)$ is used.    

\subsection{Comparison Methods under TE Schema}
Here, the comparison methods are presented in three categories: baseline, PAN winners and our TE method. The details are provided as follows.\\
\textbf{Baseline}: We connect several Machine Learning reliable classifiers widely used in the area with the seven similarity measures to set strong baselines (Table\ref{svector}). Since each example in our underlying dataset structure comprises two documents, we need to adapt it to the structure of an ordinary classifier input by converting them to one single entity. A simple direct way is to concatenate their feature vectors.  However, our experiments show it provides weak results mostly equal to the random label assignment. So, we define the \textit{summary vector} as a single unit representative of each example/problem $P=(D^S, D^T)$ by utilizing several similarity measures. The summary vector comprises a class of several metrics each measures one aspect of the closeness of the two documents ($D^S$ and $D^T$) of the pair for all underlying feature sets. For any two feature vector documents $x,y$ their summary vector is $sum(x,y)=[sim_{i}^j(x,y)] $ where $sim_{i}^j(x,y)_{ 1\leq i\leq M, 1\leq j\leq F}$ computes the $ith$ similarity metric of $M$ metrics in Table \ref{svector} under $jth$ of $F=7$ feature sets (Section \ref{setting})  between $x,y$. Then, we use a classifier including \textbf{SVM}, Gaussian Naive Bayes (\textbf{GNB}), K-Nearest Neighbor (\textbf{KNN}), Logistic Regression (\textbf{LR}), Decision Tree (\textbf{DT}) and Multi-Layer Perception (\textbf{MLP})  to predict the class label.\\
\textbf{PAN winners:} We compare our method with the top methods of PAN AV competition between 2013 and 2015.  The results of each method for one year of the competition are available and we report them here. So, our comparisons are not impacted by different parameter setting and implementation details of these methods as long as  we keep the test and training sets the same as theirs.\\
\textbf{TE methods: }We perform the Transformation Encoder (TE) on a problem $P=(D^S, D^T)$  that its documents are represented under one feature set with one feature value assignment to compute the transformation error. We then leverage the error rates taking from (at most) $F=7$ feature sets (Section \ref{setting}) of TE to form the final TE feature vector ($V$). Indeed, Each of the dimensions captures the transformation loss of one feature set. We apply the TE to both training and test data. Two well-known GNB and DT classifiers are used for verification.  We indicate them as \textbf{TE+GNB} and \textbf{TE+DT} respectively in our experiments.
\subsection{Comparison Methods under PRNN Schema} Here, the baselines are the same as the TE schema. However, we are not able to compare the PAN winner methods with our model as they have not published their code and many implementation details are left unknown to us in their reports to reimplement their methods. We call our second model \textbf{PRNN} in our reports.

 \begin{table}[t!]
 
  \def\arraystretch{1}
\centering
 
  \resizebox{0.7\textwidth}{!}{
   \begin{tabular}{llllllll}
    \toprule
    \textbf{} & \textbf{} & \multicolumn{3}{l}{\textbf{PAN2014E}} & \multicolumn{3}{l}{\textbf{PAN2014N}} \\
   
    \textbf{Category} & \textbf{Method} & \textbf{Acc.} & \textbf{AUC} & \textbf{Score} & \textbf{Acc.} & \textbf{AUC} & \textbf{Score} \\
     \midrule
    \multicolumn{1}{c}{\multirow{6}[0]{*}{\rotatebox[origin=c]{90}{\textbf{baseline}}}}  & SVM   & 0.605 & 0.309 & 0.187 & 0.57  & 0.265 & 0.151 \\
    \multicolumn{1}{c}{} & GNB   & 0.675   & 0.728 & 0.491 & 0.515 & 0.604 & 0.311 \\
    \multicolumn{1}{c}{} & LR    & 0.675  & 0.741 & 0.5 & 0.56 & 0.743 & 0.416 \\
    \multicolumn{1}{c}{} & KNN   & 0.64  & 0.689 & 0.441 & 0.55 & 0.611 & 0.336 \\
    \multicolumn{1}{c}{} & DT    & 0.64  & 0.675  & 0.432 & 0.545  & 0.545 & 0.297 \\
      \multicolumn{1}{c}{} & MLP    & \underline{0.7}  & 0.768  & 0.538 & 0.54  & 0.782 & 0.422 \\
     \midrule
    \multicolumn{1}{c}{\multirow{2}[0]{*}{\begin{sideways}\textbf{PAN}\end{sideways}}} & FCMC  & 0.58  & 0.602 & 0.349 & \textbf{0.71}  & 0.711 & 0.508 \\
    \multicolumn{1}{c}{} & OCT   &\textbf{0.71}  & 0.72  &  0.511 & 0.59  & 0.61  & 0.36 \\
     \midrule
    \multicolumn{1}{c}{\multirow{2}[0]{*}{\begin{sideways}\textbf{TE}\end{sideways}}} & TE+GNB & 0.655 & 0.676 & 0.443 & 0.685 & 0.692 & 0.474 \\
    \multicolumn{1}{c}{} & TE+DT & 0.67  & 0.675 & 0.452 & \underline{0.695} & 0.7   & 0.487 \\

     \multicolumn{8}{c}{\textbf{(A)}}
\\
     \midrule

          &       & \multicolumn{3}{l}{\textbf{PAN2013}} & \multicolumn{3}{l}{\textbf{PAN2015}} \\
         
    \textbf{Category} & \textbf{Method} & \textbf{Acc.} & \textbf{AUC} & \textbf{Score} & \textbf{Acc.} & \textbf{AUC} & \textbf{Score} \\
    \midrule
    \multicolumn{1}{c}{\multirow{6}[0]{*}{\begin{sideways}\textbf{baseline}\end{sideways}}} & SVM   & 0.633 & 0.29 & 0.184 & 0.5  & 0.215 & 0.107 \\
    \multicolumn{1}{c}{} & GNB   & 0.633   & 0.795 & 0.503 & 0.552 & 0.78 & 0.431 \\
    \multicolumn{1}{c}{} & LR    & 0.7  & 0.781 & 0.547 & 0.544 & 0.796 & 0.433 \\
    \multicolumn{1}{c}{} & KNN   & 0.633  & 0.645 & 0.409 & 0.478  & 0.464 & 0.222 \\
    \multicolumn{1}{c}{} & DT    & 0.633  & 0.621  & 0.393 & 0.558  & 0.4 & 0.223 \\
     \multicolumn{1}{c}{} & MLP    & 0.533  & 0.5  & 0.267 & 0.554  & 0.687 & 0.381 \\
     \midrule
    \multicolumn{1}{c}{\multirow{4}[0]{*}{\begin{sideways}\textbf{PAN}\end{sideways}}} & MRNN  &  -    &  -    &  -    & \textbf{0.76}  & 0.81 & 0.61 \\
    \multicolumn{1}{c}{} & Castro   &  -    &  -    &  -    & 0.694  & 0.75  & 0.52 \\
   
  &Mezaruiz    &  -    &  -    &  -    & 0.694  & 0.739  & 0.513 \\
    \multicolumn{1}{c}{} & IM    & \textbf{0.8}  & 0.792  & 0.634 &  -    &  -    &  - \\
     \midrule
        \multicolumn{1}{c}{\multirow{2}[0]{*}{\begin{sideways}\textbf{TE}\end{sideways}}}
          & TE+GNB &\textbf{0.8} & 0.835 & 0.668 & \underline{0.748} & 0.75 & 0.561 \\
          & TE+DT & \underline{0.767}  & 0.772 & 0.592 & 0.71 & 0.704   & 0.5 \\

     \multicolumn{8}{c}{\textbf{(B)}}
\\
     \bottomrule
    
    \end{tabular}%

  }
   \caption{Classification results for TE schema. All settings are kept intact according to the PAN competition. The input for the baselines are empowered by the proposed similarity vector.}
   \label{panresults}%
\end{table}%

\begin{table}[t!]
\centering
\vspace*{-\baselineskip}
    \resizebox{0.83\textwidth}{!}{%

    \begin{tabular}{lllllll}
    \toprule
    \textbf{Methods} & \textbf{MPLA*} & \textbf{Amazon} & \textbf{PAN2013} & \textbf{PAN2014E} & \textbf{PAN2014N} & \textbf{PAN2015}  \\
    \midrule
    \textbf{PRNN} & \textbf{0.703} & \textbf{0.922} & \underline{0.72}  & \textbf{0.691} & \textbf{ 0.81}  & \textbf{0.802} \\
    \midrule
    \textbf{SVM} & 0.621 & 0.818 & 0.525 & 0.659 & 0.673 & 0.628 \\
    \textbf{NB} & 0.635 & 0.741 & 0.587 & 0.652 & 0.69  & 0.728 \\
    \textbf{LR} & 0.671 & 0.839 & 0.581 & \underline{0.676} & 0.707 & 0.675 \\
    \textbf{KNN} & 0.64  & 0.831 & \textbf{0.731} & 0.656 & 0.75  & \underline{0.757} \\
    \textbf{DT} & 0.628 & 0.818 & 0.656 & 0.644 & 0.717 & 0.73 \\
    \textbf{MLP} & \underline{0.686} & \underline{0.858} & 0.65  & 0.589 & \underline{0.76}  & 0.737 \\
    \bottomrule
    \end{tabular}%
     
  }
   \caption{Classification accuracy for PRNN schema using 5 and 10-fold CV across different datasets. The input for the baselines are empowered by the proposed similarity vector.}
   \label{prnnresults}%
\end{table}%

\subsection{Results and Analysis} \label{results}
\begin{figure*}[t!]
\includegraphics[scale=.21]{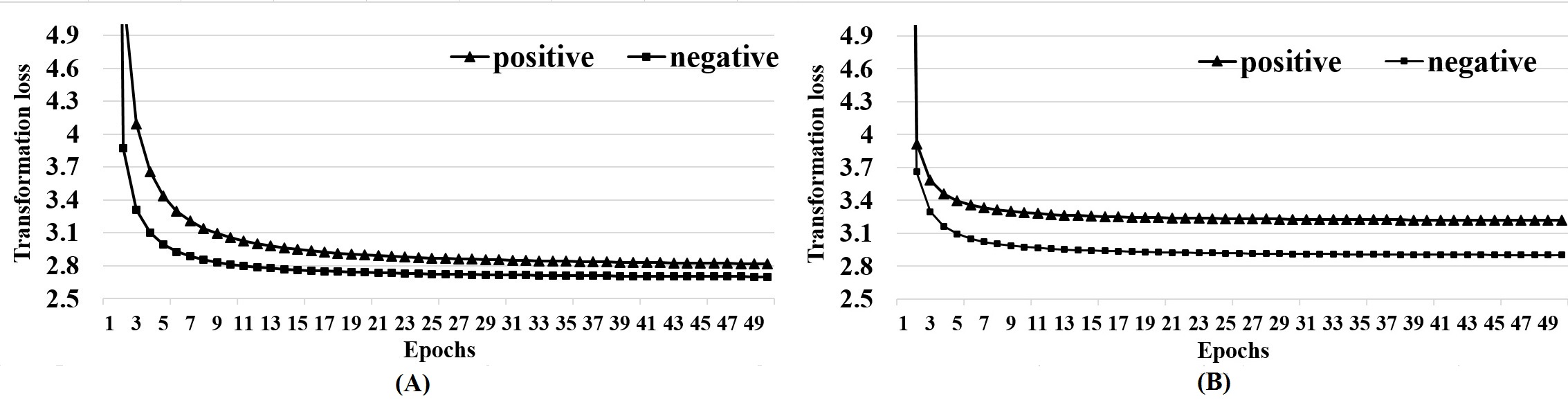}
\caption{Transformation loss for 50 epochs: averaged over all problems in the PAN2015 dataset. (A): training data (100 problems), (B): test data (500 problems), Feature set: unigram. }
\label{teloss}
\end{figure*}

\begin{figure*}[t!]
\includegraphics[scale=.22]{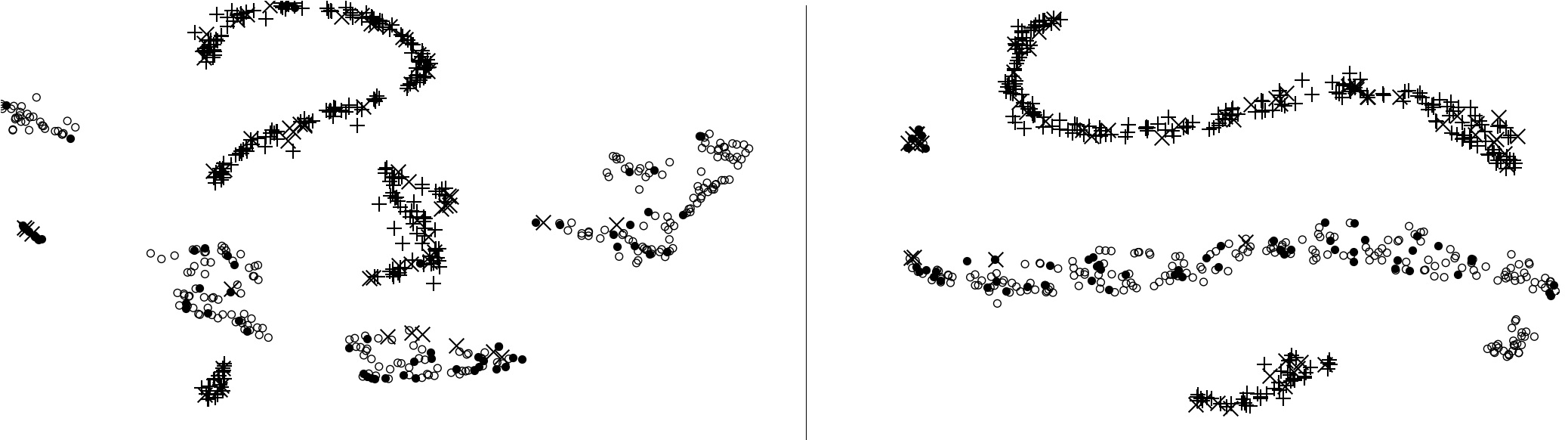}
\caption{t-SNE plot of two folds of output of the fusion layer for PAN2015 in 5-fold CV.  $+$: positive training data, $\times$: positive test data, $\circ$: negative training data, $\bullet$: negative test data }
\label{prnn figure}
\end{figure*}

\textbf{TE schema} The classification results are compared in Table~\ref{panresults}. The highest accuracy is indicated in bold and the second highest is underlined. For PAN2014N (Table~\ref{panresults}(A)) TE+GNB and TE+DT methods beat the best baseline accuracy (SVM)by more than $11$ percent and approach to the PAN winner by $1.5$ percent. For PAN2014E, TE methods overcome half of baselines and FCMC in terms of accuracy. This might be because of PAN2014E short documents compared to the PAN2014N making it harder to discriminate the employed transformation loss for positive and negative data. However, no PAN winner stays stable at the first rank in both Essays and Novels datasets. While FCMC works quite appropriately for the PAN2014N and is the winner of that year, it achieves the lowest accuracy compared to the baselines for PAN2014E. Furthermore, none of the best 2014 PAN methods (FCMC and OCT) works stably for the two genres (Novels and Essays) simultaneously. They gain the highest score in only one genre and have a large difference with the winner indicating the dependency of their method on the context while the two TE methods get close to the winners reasonably. The results for PAN2013 and PAN2015,  the dataset with the largest test set (500 pairs), are provided in (Table~ \ref{panresults}(B)). According to them, IM and TE+GNB beat the other results although most results for the PAN2013 dataset are almost close to each other in terms of accuracy. For PAN2015, Both TE+GNB and TE+DT can beat the accuracy of all the baselines. TE+GNB also shows a reasonable result while its accuracy is $1.2$ percent less than the best method (MRNN) and again stays at the second accuracy and score rank. 
Figure~\ref{teloss} shows the TE transformation loss averaged over $100$ problems of the training set and  $500$ problems of the test set of PAN2015 for both classes. The underlying feature set is unigram. The figures show that in both training and test sets the TE  loss of the two classes decreases as the transformation encoder updates its weights in each epoch. However, there is a difference between the transformation loss of the positive and negative data in both diagrams. The loss of negative data is less than the positive's. In other words, it is easier to transform one document into the other while they are a different-authorship pair (negative pair) compared to a same-authorship pair (positive pair). It makes the results of reconstruction loss to be counterintuitive. The reason is that we represent both documents of each problem under vector space model and only based on the vocabulary of the source document. So, the exclusive features of the target document, the features that only belong to the target but no to the source document,  will be filtered under this document representation model. Moreover, it is expected that the documents written by different authors have fewer features in common and have more exclusive features than the ones written by the same author. This fact makes the target document of different-authorship pair sparser than that of the same-authorship pair. And transforming the source document into a sparse document (its vector is sparse)  makes less error than to a dense document (its vector is dense). This feature differentiates the positive and negative data and exists for both training and test sets  and makes the transformation loss a distinctive feature for the verification. A more direct way to classify the documents (instead of using classifiers) is to simply thresholding the reconstruction error. However, it is  only precise  for one or two feature sets. So, we employed the NB and DT for classification.  We also tried SVM for this step but its results were not as precise as the two DT and GNB classifiers. One reason might be that SVM cannot recognize the decision boundary for very low dimensional TE error vectors (at most $7$ feature sets).\\
\textbf{PRNN schema} The comparison results are reported in Table ~\ref{prnnresults}. According to it,  PRNN  beats all baselines for all datasets except PAN2013 where it achieves the second highest accuracy.  The best accuracy belongs to the Amazon dataset where we have the largest dataset. It can be inferred that when the scale of the underlying dataset is large enough, the network learns the relation between the two language models of its given inputs well. It should be noted that for the two PAN2013 and PAN2014E even after CV the network cannot converge and the validation loss increases after each epoch. To avoid it we increase the total number of document pairs by splitting each document into two smaller ones with an equal number of sentences and making new pairs. This technique decreases the validation loss during training. However, it still suffers from lack of labeled examples and causes weakest results compared to the other larger datasets. To illustrate how PRNN  discriminate writing styles we provide the t-SNE plot of the output of the fusion layer  in a 5-fold CV classification for two folds of PAN2015 (Figure~\ref{prnn figure}). According to Figure~\ref{prnn figure}, both classes have almost similar distribution in the test and training data. But, in some rare parts, the positive and negative points are close. They are probably the portion of the data that mislead the classifier during the training step or will be misclassified in predictions.
\section{Related Work}
In majority of AV approaches the authors are known to us  and a verifier trains the language model of the future authors \cite{japkowicz1995novelty}, \cite{koppel2004authorship}, \cite{manevitz2007one},
\cite{hosseinia2017detecting}. But, in a more difficult case \textbf{no} writing samples of a questioned author are specified and they are unknown to us. No general solution has been offered for the verification problem under this assumption till 2014 \cite{koppel2014determining}. Since then, a few works can be found in the literature: Koppel and Winter \cite{koppel2014determining} propose an almost unsupervised method for the blog corpus dataset using  ``impostors" method. Optimized Classification Trees, the winner method of PAN2014 Essays dataset, optimizes a decision tree based on various types of features and different comparison methods including cosine similarity, correlation coefficient and euclidean distance \cite{frery2014ujm}. Multi-headed RNN is a character-level RNN and contains a common recurrent state among all authors with an independent softmax output per author \cite{bagnall2015author}.  Fuzzy C-Means clustering, the winner of the PAN2014 competition for novels dataset, adopts C-Means clustering and lexical features for the task \cite{modaresi2014language}. Recently, an approach based on the compression models has been evaluated on PAN datasets \cite{halvani2017usefulness}. Their method achieves promising results for the two years of PAN competitions but not for the other two datasets. Our methods is similar to these methods and considers the problems with the binary structure but we examine them on all PAN small-scale datasets as well as two large scale datasets.
 
\section{Conclusion}

Authorship verification has always been a challenging problem. It can be even more difficult when no writing samples of questioned author/authors is given. In this paper, we proposed Transformation Encoder (TE) and Parallel Recurrent Neural Network (PRNN) for small and large scale datasets. TE transforms one document of the pair into the other and observes the transformation loss as a distinctive feature for classification. PRNN investigates the difference between the language models of documents. Experiments show that TE can achieve stable results in all four PAN datasets with various size, genre and/or topics. Also, PRNN beats almost all baselines avoiding over-fitting problem by a reasonable amount of training data.
\section*{Acknowledgments}
This work is supported in part by NSF 1527364. We would like to thank the anonymous reviewers for their helpful comments.
\bibliographystyle{splncs}
\bibliography{paper}
\end{document}